\documentclass{article}

     \PassOptionsToPackage{numbers, compress}{natbib}
     \usepackage[preprint]{neurips_2020}

\usepackage[utf8]{inputenc} % allow utf-8 input
\usepackage[T1]{fontenc}    % use 8-bit T1 fonts
\usepackage[pagebackref=true,breaklinks=true,colorlinks,bookmarks=false]{hyperref}       % hyperlinks
\usepackage{url}            % simple URL typesetting
\usepackage{wrapfig,lipsum,booktabs}       % professional-quality tables
\usepackage{amsfonts}       % blackboard math symbols
\usepackage{nicefrac}       % compact symbols for 1/2, etc.
\usepackage{microtype}      % microtypography

\usepackage{amsmath}
\usepackage{amssymb}
\usepackage{graphicx}
\usepackage{subfigure}
\usepackage{multirow}

\usepackage{algorithm}      % algorithm
\usepackage{algorithmic}

\title{Gradient Boosting Neural Networks: GrowNet}

\author{%
  Sarkhan~Badirli\thanks{This work was started when authors were in Criteo AI Lab}\\
  Department of Computer Science\\
  Purdue University, Indiana, USA\\
  \texttt{sbadirli@purdue.edu}\\
  \And
  Xuanqing~Liu\\
  Department of Computer Science\\
  UCLA, California, USA\\
  \texttt{xqliu@cs.ucla.edu}
  \And
  Zhengming~Xing\\
  Linkedin, California, USA\\
  \texttt{zhxing@linkedin.com}
  \And
  Avradeep~Bhowmik\\
  Amazon, California, USA\\
  \texttt{avradeep.1@gmail.com}\\
  \And
  Khoa~Doan\\
  Department of Computer Science\\
  Virginia Tech, Virginia, USA\\
  texttt{doankhoadang@gmail.com}\\
  \And
  Sathiya S.~Keerthi\\
  Linkedin, California, USA\\
  \texttt{keselvaraj@linkedin.com }\\
}

\begin{document}

\maketitle

\begin{abstract}
A novel gradient boosting framework is proposed where shallow neural networks are employed as ``weak learners''. General loss functions are considered under this unified framework with specific examples presented for classification, regression and learning to rank.  A fully corrective step is incorporated to remedy the pitfall of greedy function approximation of classic gradient boosting decision tree. The proposed model rendered outperforming results against state-of-the-art boosting methods in all three tasks on multiple datasets. An ablation study is performed to shed light on the effect of each model components and model hyperparameters.
\end{abstract}

\section{Introduction}
%\vspace{-2pt}
%Why traditional datasets (not image and text): Concrete examples used in industry: spam emails, fraud detection, matching ads with users (Criteo)
%Concrete examples Criteo and Facebook (from XGBoost paper, 15) uses in production pipeline.

AI and machine learning pervade every aspect of modern life from email spam filtering and e-commerce, to financial security and medical diagnostics \cite{googleAI_breastcancer, Springereaav4772}. Deep learning in particular has been one of the key innovations that has truly pushed the boundary of science beyond what was considered feasible \cite{resnet, goodfellow2014generative}.

However, in spite of its seemingly limitless possibilities, both in theory as well as demonstrated practice, developing tailor-made deep neural networks for new application areas remains notoriously difficult because of its inherent complexity. Designing architectures for any given application requires immense dexterity and often a lot of luck. The lack of an established paradigm for creating an application-specific DNN presents significant challenges to practitioners, and often results in resorting to heuristics or even hacks. 

In this paper, we attempt to rectify this situation by introducing a novel paradigm that builds neural networks from the ground up layer by layer. Specifically, we use the idea of gradient boosting \cite{gbm} which has a formidable reputation in machine learning for its capacity to incrementally build sophisticated models out of simpler components, that can successfully be applied to the most complex learning tasks. Popular GBDT frameworks like XGBoost \cite{xgboost}, LightGBM \cite{lightgbm} and CatBoost \cite{catboost} use decision trees as weak learners, and combine them using a gradient boosting framework to build complex models that are widely used in both academia and industry as a reliable workhorse for common tasks in a wide variety of domains.

However, while useful in their own right, decision trees are not universally applicable, and there are many domains-- especially involving structured data-- where deep neural networks perform much better \cite{Zoph2019LearningDA, Yang2019XLNetGA, Bao2019DepthAwareVF}. In this paper, we combine the power of gradient boosting with the flexibility and versatility of neural networks and introduce a new modelling paradigm called GrowNet that can build up a DNN layer by layer. Instead of decision trees, we use shallow neural networks as our weak learners in a general gradient boosting framework that can be applied to a wide variety of tasks spanning classification, regression and ranking. We introduce further innovations like adding second order statistics to the training process, and also including a global corrective step that has been shown, both in theory \cite{rgf} and in empirical evaluation, to provide performance lift and precise fine-tuning to the specific task at hand.

Our specific contributions are summarised below:
\begin{itemize}
    \item We propose a novel approach to combine the power of gradient boosting to incrementally build complex deep neural networks out of shallow components. We introduce a versatile framework that can readily be adapted for a diverse range of machine learning tasks in a wide variety of domains.
    \item We develop an off-the-shelf optimization algorithm that is faster and easier to train than traditional deep neural networks. We introduce training innovations including second order statistics and global corrective steps that improve stability and allow finer-grained tuning of our models for specific tasks.
    \item We demonstrate the efficacy of our techniques using experimental evaluation, and show superior results on multiple real datasets in three different ML tasks: classification, regression and learning-to-rank. 
\end{itemize}

\section{Related Work}%
In this section, we briefly summarize the gradient boosting algorithms with decision trees and general boosting/ensemble methods for training neural nets.

\textbf{Gradient Boosting Algorithms.}
Gradient Boosting Machine \cite{gbm} is a function estimation method using numerical optimization in the function space. Unlike parameter estimation, function approximation cannot be solved by traditional optimization methods in Euclidean space. Decision Trees are the most common functions (predictive learners) that are used in Gradient Boosting framework. In his seminal paper, \cite{gbm} proposed Gradient Boosting Decision Trees (GBDT) where decision trees are trained in sequence and each tree is modeled by fitting negative gradients. In recent years, there have been many implementations of GBDT in machine learning literature. Among these, \cite{pGBRT} used GBDT to perform learning to rank, \cite{add_logistic_reg} did classification and \cite{xgboost, lightgbm} generalized GBDT for multi-tasking purposes. In particular, scalable framework of \cite{xgboost}  made it possible for data scientists to achieve state-of-the-art results on various industry related machine learning problems. For that reason, we take XGBoost \cite{xgboost} as our baseline. Unlike these GBDT methods, we propose gradient boosting neural network where we train gradient boosting with shallow neural nets. Using neural nets as base learners also gives our method an edge over GBDT models, where we can  correct each previous model after adding the new one, referred to as ``corrective step'', in addition to the ability to propagate information from the previous predictors to the next ones.

\textbf{Boosted Neural Nets.}
\iffalse
\#\#\# This part is just the flow of the section Boosted  neural nets in related work.
\begin{enumerate}
    \item cascade paper for growing NNs \cite{cascade_fahlman}
    \item Ada-Boost \& ensemble based NN learning \cite{boosting_NN, Adaboost_artificial_NN, training_adaboost_NN, ensemble_NN, ensembling_NN}
    \item Recent attempts to decompose Deep NN as ensemble of low-biased shallow nets \cite{ResNet_as_ensemble, modern_small_NN}
    \item Gradient Boosting CNN frameworks \cite{gb_random_cnn, boosted_CNN}: \cite{gb_random_cnn} is closest one. They have done Gradient boossting CNN but just using one custom loss function and tested the model on 2 remote sensing problem. 
    \item Hybrid models: features from DNN and classification from XGBoost \cite{hybrid_gbdt_NN} -- I didn't add this part as their approach is more of a hack than a rigorous approach.
\end{enumerate}
\fi
Although weak learners, like decision trees, are popular in boosting and ensemble methods, there have been a substantial work done on combining neural nets with boosting/ensemble methods for better performance over single large/deep neural networks. 
The idea of considering shallow neural nets as weak learners and constructively combining them started with \cite{cascade_fahlman}. In their pioneering work, fully connected, multi-layer perceptrons are  trained in a layer-by-layer fashion and added to get a cascade-structured neural net. Their model is not exactly a boosting model as the final model is a single, multi-layer neural network. 

In 1990's, ensemble of neural networks got popular as ensemble methods helped to significantly improve the generalization ability of neural nets. Nevertheless, these methods were simply either majority voting \cite{ensemble_NN} for classification tasks, simple averaging \cite{ensemble_NN2} or weighted averaging \cite{ensemble_NN3} for regression tasks. After the introduction of adaptive boosting ($Adaboost$) algorithm \cite{adaboost}, \cite{training_adaboost_NN} investigated boosting with multi-layer neural networks for a character recognition task and achieved a remarkable performance improvement. They extended the work to traditional machine learning tasks with variations of Adaboost methods where different weighting schemes are explored \cite{boosting_NN}. The adaptive boosting can be seen as a specific version of the gradient boosting algorithm where a simple exponential loss function is used \cite{add_logistic_reg}. %(I think we need to put some sentences here to mention the differences between these methods and ours).

In early 2000's, \cite{hinton_greedy_layerwise_DBN}
introduced greedy layer-wise unsupervised training for Deep Belief Nets (DBN). DBN is built upon a layer at a time by utilizing Gibbs sampling to obtain the estimator of the gradient on the log-likelihood of Restricted Boltzmann Machines (RBM) in each layer. The authors of \cite{bengio_greedy_layerwise} expounded this work for continuous inputs and explained its success on attaining high quality features from image data. They concluded that unsupervised training helped model training by initializing RBM weights in a region close to a good local minimum.

Most recently, AdaNet~\cite{adanet} was proposed to adaptively built Neural Network (NN) layer by layer from a singe layer NN to perform image classification task. Beside  learning network weights, AdaNet adjusts the network structure and its growth procedure is reinforced by a  theoretical justification. AdaNet optimizes over a generalization bound that consists of empirical risk and complexity of the architecture. Coordinate descend approach is applied to the objective function, and heuristic search (weak learning algorithm) is performed to obtain $\delta-optimal$ coordinates. Although the learning process is boosting-style, the final model is a single NN whose final output layer is connected to all lower layers. Unlike AdaNet, we train each weak learner in a gradient boosting style, resulting in less entangled training. The final prediction is the weighted sum of all weak learners' output. Our method also renders a unified platform to perform various ML tasks.

In recent years, a few work have been done to explain the success of deep residual neural networks \cite{resnet} with hundreds of layers by showing that they can be decomposed into a collection of many subnetworks.
The work in \cite{boostresnet} extends AdaNet to specifically focus on ResNet architecture~\cite{resnet} to provide a new training algorithm for ResNet. The authors of \cite{ResNet_as_ensemble}, meanwhile, argue that these deeper layers might serve as a bagging mechanism in a similar spirit to random forest classifier. These studies challenge the common belief that neural networks are too strong to serve as weak learners for boosting methods.

% short paths rather than a very long one. These short paths do not show strong dependence and they exhibits ensemble-like behavior \cite{ResNet_as_ensemble}. Authors of the work argue that these deeper layers might serve as a bagging mechanism in a similar spirit to random forest classifier. \cite{modern_small_NN} even argues that slightly tuning deep neural networks one can get on par results on many real-world datasets from the UCI Machine Learning Repository. Furthermore, they empirically demonstrate  that neural networks can be decomposed into an ensemble of sub-networks each of which achieves low training errors.

\begin{figure*}[t]
\center
  \includegraphics[width=0.95\textwidth]{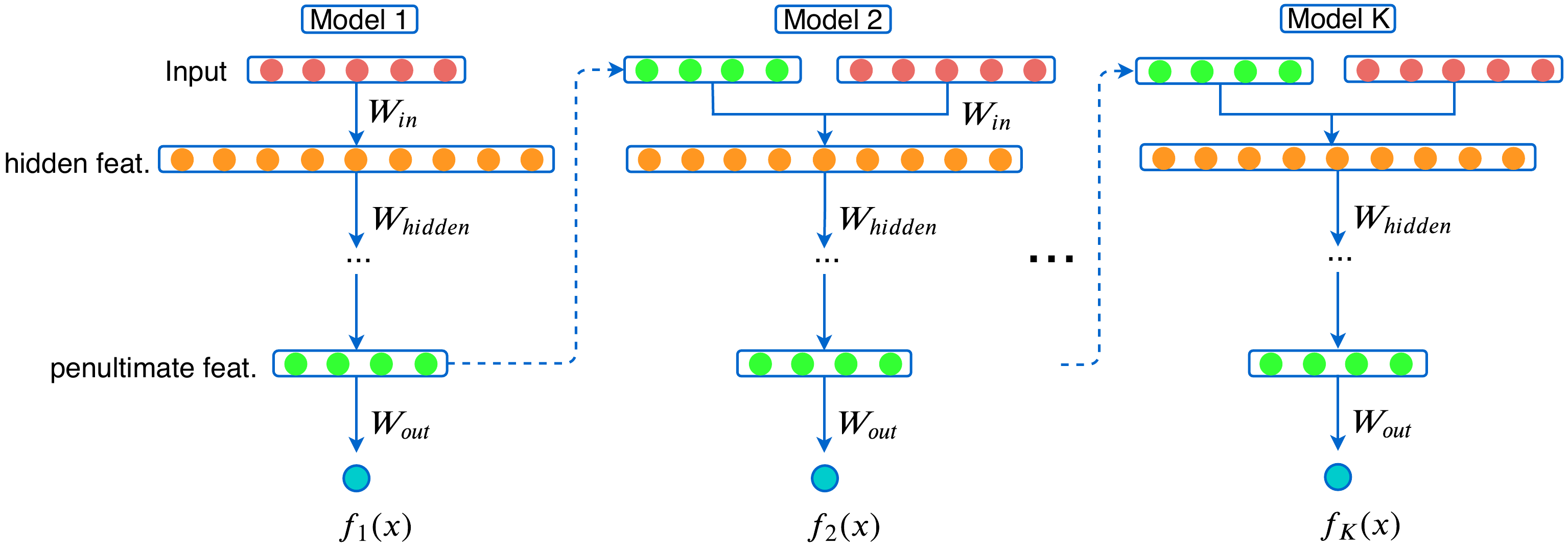}
  \caption{GrowNet architecture. After the first weak learner, each predictor is trained on combined features from original input and penultimate layer features from previous weak learner. The final output is the weighted sum of outputs from all predictors,  $\sum_{k=1}^{k=K}\alpha_k f_k(x)$. Here Model K means weak learner K.}
  \label{fig:stacked_en}
\vspace{-15pt}
\end{figure*}

%\fi

% \begin{algorithm}[tb]
%     \caption{Full GrowNet training}
%     \label{algo:main}
% \begin{algorithmic}
%     \STATE \textbf{Input:} $f_0(x) = log(\frac{n_+}{n_-})$, $\alpha_0$, Training data $\mathcal{D}_{tr}$\\
%     \STATE \textbf{Output:} GrowNet $\mathcal{E}$\\
% \FOR{$k = 1$ {\bfseries to} $M$}
%     \STATE \# Individual model training
%     \STATE Initialize model $f_k(x)$
%     \STATE Calc. $1^{st}$ order grad.: $g_i = \partial_{\hat{y}_{i}^{(k-1)}} l (y_i, \hat{y}_i^{(k-1)}),$ $\forall x_i \in \mathcal{D}_{tr}$ 
%     \STATE Calc. $2^{nd}$ order grad.: $h_i = \partial^{2}_{\hat{y}_{i}^{(k-1)}} l (y_i, \hat{y}_i^{(k-1)}),$ $\forall x_i \in \mathcal{D}_{tr}$ \\
%     \STATE Train $f_k(\cdot)$ by least square regression on $\{x_i, -g_i / h_i\}$ \\
%     \STATE Add the model $f_k(x)$ into the GrowNet $\mathcal{E}$ \\

%     \STATE \# Corrective step\\
%     \FOR{$epoch = 1$ {\bfseries to} $T$}
%         \STATE Calc. GrowNet output: $\hat{y}_i^{(k)} = \sum_{m=0}^{k} \alpha_m f_m (x_i)$, $\forall x_i \in \mathcal{D}_{tr}$ 
%         \STATE Calculate Loss from GrowNet: $\mathcal{L} = \frac{1}{n} \sum_i^{n} l (y_i, \hat{y}_i^{(k)})$
%         \STATE Update model $f_i$ parameters through back-propagation  $\forall i \in \{1, ...k\}$\\
%         \STATE Update step size $\alpha_k$ through back-propagation
%     \ENDFOR
% \ENDFOR
% \end{algorithmic}
% \end{algorithm}

\vspace{-3pt}
\section{Model}\label{model}
\vspace{-3pt}
%Neural networks, in particular CNNs and RNNs, have dominated the computer vision and natural language processing tasks. For traditional machine learning tasks, however, gradient boosting decision tree (GBDT) \cite{gbm}  based methods  and their derivatives, XGBoost \cite{xgboost}, LightGBM \cite{lightgbm}, overshadow the neural networks. In this paper, we propose  gradient boosting neural network (GrowNet) that achieves competitive results on traditional ML tasks of classification, regression and learning to rank.
In this section, we first describe the basic framework of GrowNet for general loss functions, and then we show how the corrective step is incorporated. 
The key idea in gradient boosting is to take simple, lower-order models as weak learners and use them as fundamental building blocks to build a powerful, higher-order model by sequential boosting using first or second order gradient statistics. We use shallow neural networks (e.g., with one or two hidden layers) as weak learners in this paper. As each boosting step, we augment the original input features with the output from the penultimate layer of the current iteration (see Figure \ref{fig:stacked_en}). This augmented feature-set is then fed as input to train the next weak learner via a boosting mechanism using the current residuals. The final output of the model is a weighted combination of scores from all these sequentially trained models.

%=====

\subsection{Gradient Boosting Neural Network: GrowNet}

Let us assume a dataset with $n$ samples in $d$ dimensional feature space $\mathcal{D} = \{(\boldsymbol{x}_i, y_i)_{i=1}^{n} | \boldsymbol{x}_i \in \mathbb{R}^d, y_i \in \mathbb{R}\}$. GrowNet  uses $K$ additive functions to predict the output,
\begin{equation}
\hat{y}_i = \mathcal{E} (\boldsymbol{x}_i) = \sum_{k=0}^{K} \alpha_k f_k (\boldsymbol{x}_i), f_k \in \mathcal{F}
\end{equation}

where $\mathcal{F}$ is the space of multilayer perceptrons and $\alpha_k$ is the step size (boost rate). Each function $f_k$ represents an independent, shallow neural network with a linear layer as an output layer. For a given sample $\boldsymbol{x}$, the model calculates the prediction as a weighted sum of $f_k$'s in GrowNet.

Let $l$ be any differentiable convex loss function. Our objective is to learn a set of functions (shallow neural networks) that minimize the following equation: $\mathcal{L}(\mathcal{E}) = \sum_{i=0}^{n} l(y_i, \hat{y}_i).$
% \begin{equation} \label{eq:generalLoss}
% \mathcal{L}(\mathcal{E}) = \sum_{i=0}^{n} l(y_i, \hat{y}_i) 
% \end{equation}

We may further add regularization terms to penalize the model complexity but it is omitted for simplicity in this work. As the objective we are optimizing is over the functions and not on the parameters, traditional optimization techniques will not work here. Analogous to GBDT \cite{gbm}, the model is trained in an additive manner.

Let $\hat{y}^{(t-1)}_i = \sum_{k=0}^{t-1} \alpha_k f_k(\boldsymbol{x}_i)$ be the output of GrowNet at stage $t-1$ for the sample $\boldsymbol{x}_i$.
We greedily seek the next weak learner  $f_{t}(\textbf{x})$ that will minimize the loss at stage $t$ which can be summized as,
\begin{equation}
\mathcal{L}^{(t)} = \sum_{i=0}^{n} l(y_i, \hat{y}_i^{(t-1)}+\alpha_t f_{t}(\textbf{x}_i))
\end{equation}
In addition, Taylor expansion of the loss function $l$ was adopted to ease the computational complexity. As second-order optimization techniques are proven to be superior to first-order ones and require less steps to converge, we train models with Newton-Raphson steps. Consequently, regardless of the ML task, individual model parameters are optimized by running regression on the second order gradients of the GrowNet's outputs. Objective function for the weak learner $f_t$ can be simplified as follows,
\begin{equation}
    \mathcal{L}^{(t)} = \sum_{i=0}^{n} h_i (\tilde{y}_i - \alpha_t f_t(\boldsymbol{x}_i))^{2}
\end{equation}
where $\tilde{y}_i = -g_{i}/h_{i}$, and $g_{i}$ \& $h_{i}$ are the first and second order gradients of the objective function $l$ at $\boldsymbol{x}_i$, w.r.t. $\hat{y}_i^{(t-1)}$. 
(See pseudo-code in part 1 of Algorithm 1 from supplementary material.)
%The pseudo-code of the procedure is explained in Individual model training part of algorithm 1 in supplementary material.%\ref{algo:main} .

\subsection{Corrective Step (C/S)}
In a traditional boosting framework, each weak learner is greedily learned. This means that only the parameters of $t^{th}$ weak learner are updated at boosting step $t$ where all the parameters of previous $t-1$ weak learners remain unchanged. The myopic learning procedures may cause the model to get stuck in a local minima, and a fixed boosting rate $\alpha_{k}$ aggravates the issue \cite{gbm}. Therefore, we implemented a corrective step to address this problem. In the corrective step, instead of fixing the previous $t-1$ weak learners, we allow update of the parameters of the previous $t-1$ weak learners through back-propagation. Moreover, we incorporated the boosting rate $\alpha_k$ into parameters of the model and it is automatically updated through the corrective step. Beyond getting better performance, this move saves us from tuning a delicate parameter. C/S can also be interpreted as a regularizer to mitigate the correlation among weak learners, as during corrective step, the main training objective becomes task specific loss function on just original inputs. The usefulness of this step is empirically and theoretically investigated in \cite{rgf} for gradient boosting decision tree models. Our experiments in the ablation study~\ref{cs} further validate the necessity of c/s in our model as well. The corrective step is summarized in the second part of Algorithm 1 in the supplementary material.
\vspace{-3pt}
\section{Applications}
\vspace{-3pt}
In this section, we show how GrowNet can be
adapted for regression, classification and learning to rank problems.

\textbf{GrowNet for Regression.}
We employ mean squared error (MSE) loss function for the regression task. Let us assume $l$ is the MSE loss; then we can easily obtain $\tilde{y}_i$, first order, and second order statistics at stage $t$, as follows:
\begin{align*}
    g_i = 2(\hat{y}_i^{(t-1)} - y_i), \quad h_i = 2 \implies \tilde{y}_i =  y_i - \hat{y}_i^{(t-1)}
\end{align*}
We train next weak learner $f_{t}$ by least square regression on $\{\boldsymbol{x}_i, \tilde{y}_i\}$ for $i=1, 2, ..., n$. In the corrective step, all model parameters in GrowNet are updated again using the MSE loss.

\textbf{GrowNet for Classification.}
 For the illustration purposes, let us consider the binary cross entropy loss function; however, note that any differentiable loss function can be used. Choosing labels $y_i \in \{-1, +1\}$ (this notation has an advantage of $y_i^{2}=1$, which will be used in the derivation), the first and second order gradients, $g_i$ and $h_i$ , respectively, at stage $t$ can be written as follows,
\begin{align*}
    g_i = \frac{-2y_i}{1 + e^{2y_i \hat{y}_i^{(t-1)}}}, \quad h_i = \frac{4 y_i^{2} e^{2y_i \hat{y}_i^{(t-1)}}}{(1 + e^{2y_i \hat{y}_i^{(t-1)}})^2} \implies
    \tilde{y}_i = -g_i/h_i = y_i (1 + e^{-2 y_i \hat{y}_i^{(t-1)}})/2
\end{align*}
% \begin{align*}
%     g_i & = \frac{-2y_i}{1 + e^{2y_i \hat{y}_i^{(t-1)}}}, \quad h_i = \frac{4 y_i^{2} e^{2y_i \hat{y}_i^{(t-1)}}}{(1 + e^{2y_i \hat{y}_i^{(t-1)}})^2}\\
%     \tilde{y}_i & = -g_i/h_i = y_i (1 + e^{-2 y_i \hat{y}_i^{(t-1)}})/2
% \end{align*}
The next weak learner $f_{t}$ is fitted by least square regression using second order gradient statistics on $\{x_i, \tilde{y}_i\}$. In the corrective step, parameters of all the added predictive functions are updated by retraining the whole model using the binary cross entropy loss. This step slightly corrects the weights according to the main objective function of the task at hand, i.e. classification in this case.

\textbf{GrowNet for Learning to Rank.}
In this part, we demonstrate how the model is adapted to learning to rank (L2R)  with a pairwise loss. In the L2R framework, there are queries and documents associated with each query. A document can be associated with many different queries. Then for each query, the associated documents have relevance scores. Assume for a given query, a pair of documents $U_i$ and $U_j$ is chosen. Assume further that we have a feature vector for these documents, $\boldsymbol{x}_i$ and $\boldsymbol{x}_j$. Let $\hat{y}_i$ and  $\hat{y}_j$ denote the output of the model  for samples $\boldsymbol{x}_i$ and $\boldsymbol{x}_j$ respectively. 
According to \cite{l2r_overview}, a common  pairwise loss  for a given query can be formulated as follows,
$$l(\hat{y}_i, \hat{y}_j) = \frac{1}{2} (1 - S_{ij})\sigma_0(\hat{y}_i - \hat{y}_j) + log(1 + e^{-\sigma_0 (\hat{y}_i - \hat{y}_j)})$$
where $S_{ij} \in \{0, -1, +1\}$ denotes the documents' relevance difference; it is $1$ if the $U_i$ has a relevance score greater than $U_j$, $-1$ vice-versa and $0$ if both document have been labeled with the same relevance score. $\sigma_0$ is the sigmoid function.
Note that the cost function $l$ is symmetric and  its gradients can be easily computed as follows (for the details, readers can refer to \cite{l2r_overview}),
\begin{align}
    \partial_{\hat{y}_i} l(\hat{y}_i, \hat{y}_j) & = \sigma_0 (\frac{1}{2} (1 - S_{ij}) - \frac{1}{1 + e^{\sigma_0 (\hat{y}_i - \hat{y}_j)}})\nonumber = - \partial_{\hat{y}_j} l(\hat{y}_i, \hat{y}_j ) \nonumber\\
    \partial_{\hat{y}_i}^{2} l(\hat{y}_i, \hat{y}_j) & = \sigma_0^{2} (\frac{1}{1 + e^{\sigma_0 (\hat{y}_i - \hat{y}_j)}}) (1 - \frac{1}{1 + e^{\sigma_0 (\hat{y}_i - \hat{y}_j)}}) \nonumber
\end{align}
where $I$ denotes the set of pairs of indices $\{i, j\}$, for which $U_i$ is desired to be ranked differently from $U_j$ for a given query.
Then for a particular document $U_i$, the loss function and its first and second order statistics can be derived as follows,
\begin{align}
    l & = \sum_{j:\{i, j\} \in I} l(\hat{y}_i, \hat{y}_j) + \sum_{j:\{j, i\} \in I} l(\hat{y}_i, \hat{y}_j)\nonumber \\ \nonumber
     g_i & = \sum_{j:\{i, j\} \in I} \partial_{\hat{y}_i} l(\hat{y}_i, \hat{y}_j) - \sum_{j:\{j, i\} \in I} \partial_{\hat{y}_i} l(\hat{y}_i, \hat{y}_j), \quad
    h_i & = \sum_{j:\{i, j\} \in I} \partial_{\hat{y}_i}^{2} l(\hat{y}_i, \hat{y}_j) - \sum_{j:\{j, i\} \in I} \partial_{\hat{y}_i}^{2} l(\hat{y}_i, \hat{y}_j) \nonumber
\end{align}
\section{Experiments}
\textbf{Experiment Setup.} All predictive functions added to the model are multilayer perceptrons with two hidden layers. We generally set the number of  hidden layer units  to roughly half of or equal to the input feature dimension. More hidden layers degraded the performance as the model starts overfitting. $40$ additive functions were employed in the experiments for all three tasks and the number of weak learners in test time is chosen by the validation results. Boosting rate is initially set to $1$ and  automatically adjusted during the corrective step.

We trained each predictive function for just $1$ epoch and the entire model is also trained for $1$ epoch during the corrective step by stochastic gradient descent with Adam optimizer. The number of epochs is increased to 2 for the ranking task. We also employed $2D$ batch normalization on the hidden layers. %Increasing the epoch number more than that does not contribute to the performance and higher numbers cause overfitting. 
%We designed 2 versions of the model: simple and stacked one. In the simple version input feature for each additive function is kept same yet in the stacked model, as the name suggests, we combined original input features with penultimate layer features from the previous predictor (see Figure \ref{fig:stacked_en}).
We compared the model performance with XGBoost since similar results are obtain with LightGBM or CatBoost and with AdaNet. Tuning and model details of all 3 methods are provided in supplementary material. 

% \begin{table}[t]
% \caption{Datasets used in the experiments and their brief description. The second and third columns marked as N, M represent number of samples and feature dimension of the dataset, respectively.}
% \label{tab: datasets}
% \begin{center}
% \begin{tabular}{|l|c|c|c|}
% \hline
%     Dataset    & N & M & Task\\ \hline
% Higgs Bozon & $10M$ & $28$ & Binary classification  \\ \hline
% %Criteo ?    & & &  \\\hline
% Slice localization & $53K$ & $384$ &  Regression \\\hline
% Year prediction & $515K$ & $90$ &  Regression \\ \hline
% Yahoo LTRC & $473K$ & $700$ & Learning to rank \\ \hline
% MSLR-WEB 10K & $1.2M$ & $136$ & Learning to rank \\ \hline
% \end{tabular}
% \end{center}
% \end{table}

\textbf{Datasets.} We evaluate our model on 5 datasets from 3 different tasks. 
Higgs Bozon dataset is used for classification. Higgs data is created using Monte Carlo simulations on high energy physics events. %It is a binary event classification data with 28 attributes.

To perform regression, 2 datasets from UCI machine learning repository are selected. The first one is Computed Tomography (CT) slice localization data where the aim is to retrieve the location of CT slices on axial axis. %The data was constructed from a set of $53,500$ CT images that were taken from 74 different patients (43 male, 31 female). 
The second regression dataset is YearPredictionMSD, a subset of Million Song dataset. The goal is to predict the release year of a song from its audio features. %Songs are mostly western, commercial tracks ranging from 1922 to 2011, with a peak in the year 2000s.

We choose Yahoo LTR dataset \cite{yahoo_ltrc} for the learning to rank task as it is a well-known benchmark dataset and also used in XGBoost's paper. The dataset has $20K$ queries each associated with approximately $22$ documents. Train-test split from the original paper is preserved. The second benchmark ranking dataset we used is MSLR-WEB $10K$ in which there are $10K$ queries, each corresponding to list of $100-200$ documents. Detailed statistics of each dataset can be found in the supplementary material.
\begin{table*}[t]
\center
\caption{L2R results in Normalized Discounted Cumulative Gain for top 5 and 10 queries (NDCG@5 \& 10), on Microsoft Learning to Rank with 10K queries and Yahoo LTR datasets. GrowNet results are average of 5 iterations and the values in the parenthesis represents the standard deviation. }
\resizebox{\textwidth}{!}{
\begin{tabular}{|l|cc|cc|}
\hline
        & \multicolumn{2}{c}{MSLR-WEB 10K} & \multicolumn{2}{c}{Yahoo LTR} \\ \hline
        & NDCG@5 & NDCG@10 & NDCG@5 & NDCG@10 \\ \hline
XGBoost & $0.4677 (0.0287)$ & $0.4858 (0.0245)$   &  $0.7618$ & $0.7913$             \\ \hline
GrowNet (pairwise loss)    & $\boldsymbol{0.5106} (0.0011)$  & $\boldsymbol{0.5203} (0.0015)$ & $\boldsymbol{0.7726} (0.0006)$ & $\boldsymbol{0.8101} (0.0003)$  \\\hline
GrowNet (Gen. I div. loss)  & $0.5044 (0.0072)$ & $0.5137 (0.0070)$ & $0.7713 (0.0006)$ & $0.8088 (0.0005)$   \\\hline
\end{tabular}
\label{tab: l2r_results}
}
\vspace{-10pt}
\end{table*}

\subsection{Results}
\begin{wraptable}{r}{0.5\textwidth}
\vspace{-20pt}
\begin{minipage}{0.5\textwidth}
\caption{Regression results in root mean square error (RMSE). GrowNet results are average of 5 iterations and the values in the parenthesis represent standard deviation.}
\label{tab: reg_results}
\begin{center}

\begin{tabular}{|l|c|c|}
\hline
        & Music Year Pred. & Slice Localz. \\ \hline
XGBoost & $8.9301$                & $6.6744$             \\ \hline
AdaNet  & $12.1778$ & $5.3824$ \\ \hline
GrowNet    & $\boldsymbol{8.8156}$  (0.0061)      & $\boldsymbol{5.3112}$  (0.3512)  \\\hline
\end{tabular}

\end{center}
\end{minipage}
%\vspace{15pt}
\begin{minipage}{0.5\textwidth} %\begin{table}[t]
\caption{Classification results, in AUC, on Higgs bozon dataset. For our model, we preset 3 different results: using all the data, $10\%$ of the data ($1M$), and $1\%$ of the data ($100K$).}
\label{tab: cls_results}
\begin{center}
\begin{tabular}{|l|c|}
\hline
XGBoost & $0.8304$           \\ \hline
GrowNet (all data)    & $\boldsymbol{0.8510}$      \\\hline
GrowNet (data sampling$=10\%$)    & $0.8439$     \\\hline
GrowNet (data sampling$=1\%$)    & $0.8180$  \\\hline
\end{tabular}
\end{center}
\end{minipage}
\vspace{-10pt}
\end{wraptable}
\textbf{Regression.} Table \ref{tab: reg_results} reports regression performance on two UCI datasets. GrowNet outperforms both methods on Music dataset where AdaNet delivers the worse result. On  CT slice localization dataset, our model obtains on par results with Adanet and displays $21\%$ decrease in RMSE compared to XGBoost.

\textbf{Classification.} To make a fair comparison with XGBoost, we tested our model on Higgs bozon dataset as it is used in XGBoost's paper \cite{xgboost}. Classification results are presented in the Table \ref{tab: cls_results}.  GrowNet clearly outperforms XGBoost using all the data. Subsampling $10\%$ of the data for training each weak learner also renders better performance. We used 30 weak learners (multilayer perceptrons with two hidden layers of $16$ units) and the number of weak learners to be used at test time is chosen by validation results. In all 3 experiments, this number was 30.

\textbf{Learning to Rank.} Ranking experiment results  on Yahoo and MSLR datasets are presented in Table \ref{tab: l2r_results}. We evaluated GrowNet with 2 different loss functions, namely pairwise loss and generalized I-divergence loss. In both scenarios, GrowNet outperforms XGBoost on both datasets; in particular, it delivers $6-8\%$ increase on Microsoft data in  NDCG@5 and NDCG@10. For our model to achieve these results, 30 weak learners were enough.

\section{Ablation study}
\begin{wrapfigure}{r}{0.50\textwidth}
\vspace{-10pt}
\begin{centering}
\includegraphics[width=0.49\textwidth]{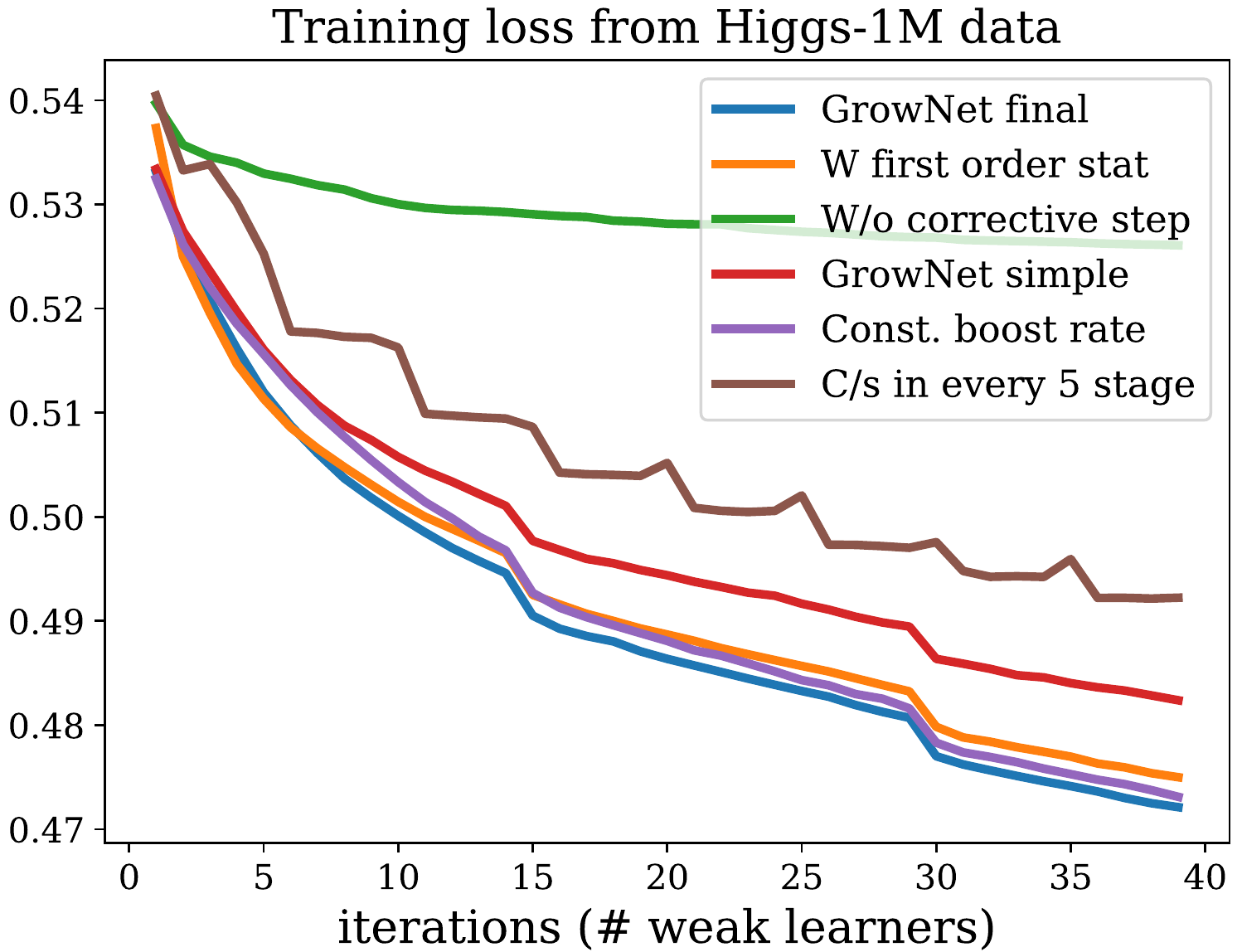}
\end{centering}
\vspace{-5pt}
\caption{Classification training losses}
\label{tr_losses}
\vspace{-20pt}
\end{wrapfigure}
We investigated different components of GrowNet. We picked 2 datasets for these experiments: Higgs and Microsoft. For Higgs dataset, we randomly selected $1M$ points for training and $5\%$ of the remaining as the validation set. The original test data was used as the test set. For Microsoft dataset, we used Fold 1 and the original split was preserved.  In all upcoming experiments, only the component that is being analyzed, is altered while the rest of the parameters remain unchanged. All ablation experiments are reported in Table \ref{tab: ablation}, and the third column (GrowNet) represents the results from final version of our model on these datasets.

\begin{table*}[t]
\caption{Ablation study experiment results on Higgs $1M$ and Microsoft (Fold 1) datasets. All models have two-layer shallow networks as weak learners. Hidden layer dimension is 16 for classification and 64 for ranking task. The third column is the final GrowNet model that all different versions are compared against. Reported results are AUC scores for classification and NDCG for ranking.}
\label{tab: ablation}
\center
\resizebox{\textwidth}{!}{
\begin{tabular}{l|l|c|c|c|c|c|c}
%\hline
Datasets & Eval. metric & GrowNet & $1^{st}$ order grad. & Constant $\alpha_t$ & Simple version & No C/S & C/S in every 5 stage \\ \hline
Higgs $1M$ & AUC &$\boldsymbol{0.8401}$ & $0.8363$ & $0.8397$ & $0.8326$ & $0.8093$ & $0.8315$\\ \hline
\multirow{2}{*}{MSLR Fold1} & NDCG@5 & $\boldsymbol{0.5106}$ & $0.5001$ & $0.5020$& $0.4836$ & $0.4743$ & $0.4881$\\ 
& NDCG@10 & $\boldsymbol{0.5195}$ & $0.5104$ &$0.5115$& $0.4972$ & $0.4872$ & $0.4998$\\ %\hline
\end{tabular}
}
\end{table*}

% \begin{figure*}[ht]
% \begin{centering}
% \subfigure[Classification on Higgs 1M]{\includegraphics[width=0.49\textwidth, trim=0 0 10 0, clip]{figures/all_exp_higgs_trloss_new.pdf}}
% \subfigure[]{\includegraphics[height=4.75cm, width=0.24\textwidth]{figures/hu_exp_higgs_trloss.pdf}}%\label{fig:hu_exp}
% \subfigure[]{\includegraphics[height=4.25cm, width=0.24\textwidth]{figures/dynamic_boost_rate.pdf}}%\label{fig:hu_exp}
% \end{centering}
% %\vspace{3mm}
% \caption{Effect of hidden layers and units on classification performance (AUC).}
% \label{fig:hl_hu_exp}
% \label{fig:1vs2}
% \end{figure*}

\subsection{Stacked versus simple version}
As seen in Figure \ref{fig:stacked_en}, every weak learner except the first one is trained on the combined features of the original input and penultimate layer's features from previous predictive function. It is worth to note that the input dimension does not grow by iteration; indeed, it is always the dimension of hidden layer plus the dimension of of original input. This idea of stacked features has a weak resemblance to auto-context \cite{auto_context} in literature, where the authors utilized the direct output of the classifier, along with the original inputs, to boost the image segmentation performance. The work in \cite{kernelboost} extended this idea to not only use the output of the classifier, probabilities, but also the raw prediction image itself. Our model is significantly different from these methods, as we do not simply use the previous model’s output but more expressive representation at the  penultimate layer. These features leverage our model by propagating more complex information from previous model to the new one. To test the advantage of this stacked model, we compared the proposed model against its simpler version in which the original input features are used for all learners. The sixth column in Table \ref{tab: ablation} presents the results from the simpler version. In both tasks, the stacked model outperforms the simpler version; especially, the difference is noteworthy in the ranking task. Training loss in Figure \ref{tr_losses} also supports the information gain while the stacked version is utilized. Unlike tree boosting methods, our model makes this architecture possible through its flexible weak learners.

\subsection{Analyzing corrective step}\label{cs}

Among all components of the model, the corrective step is presumably the most vital one. In this step, the parameters of all weak learners, that are added to the model, are updated by training the whole model on the original inputs without the penultimate layer features. The loss function used in this step is a task specific one. This procedure allows the model to rectify the parameters to specifically better accommodate the task at hand rather than fitting negative gradients. C/S also alleviates the potential correlation among weak learners. Moreover, within this step, we incorporated the boosting rate $\alpha_t$ and it is automatically adjusted without requiring any tuning. The last two columns of Table \ref{tab: ablation} present the classification and learning to rank results from GrowNet without using any corrective step and using a corrective step in every 5 stages, respectively. The performance severely degraded in the former one, and the model hardly learned any  information after a couple of predictive functions added. The flat training  loss in Figure \ref{tr_losses} confirms this phenomenon as well. Running the corrective step in every 5 steps rendered much better performance, yet was not as good as GrowNet's results. The stair-like loss curve in the Figure \ref{tr_losses} evidently displays the influence of the corrective step on model training.
%\vspace{-0.4cm}

\begin{wrapfigure}{r}{5cm}
\vspace{-10pt}
\includegraphics[width=0.35\textwidth]{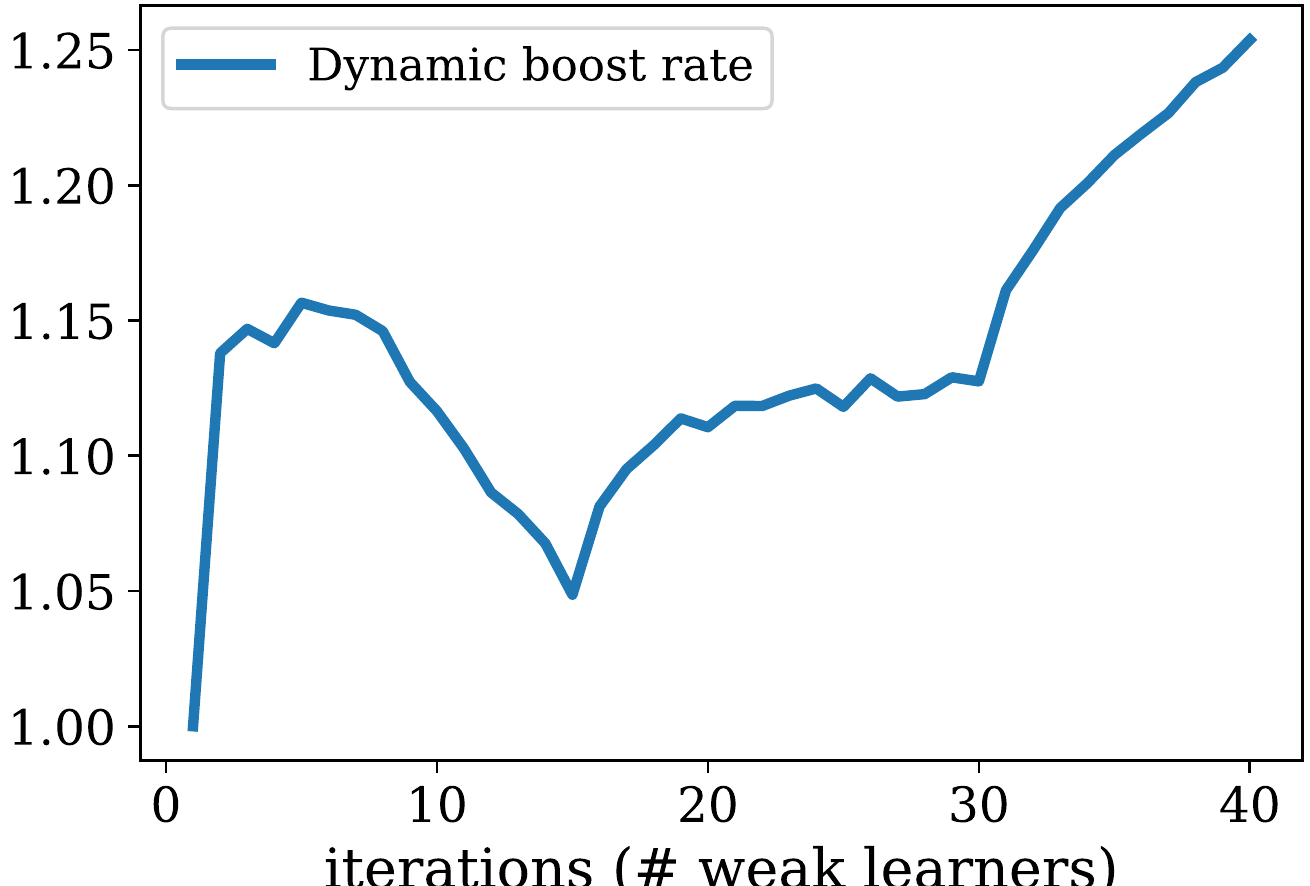}
\caption{Boosting rate evolution}
\label{fig:dynamic_br}
\vspace{-20pt}
\end{wrapfigure}
\textbf{Dynamic boost rate.} Within the corrective step, we are able to dynamically update the boost rate $\alpha_t$ (at stage $t$). Taking this measure saved us from tuning one more parameter as well as yielded a mild performance increase in all tasks. Moreover, the model obtained better training loss convergence, compared to the fixed boost rate version (see Fig. \ref{tr_losses}). In our setup, starting with $\alpha_0=1$, the boost rate is automatically updated each time the corrective step is executed (see Fig. \ref{fig:dynamic_br}) . Results of the best constant boost rate ($\alpha_t=0.1$), coarsely tuned in a set of $\{0.01, 0.1, 1\}$,  are reported in fifth column of Table \ref{tab: ablation}. 
\subsection{First order statistics vs second order statistics}
In this experiment, we explored the impact of first and second order statistics on model performance as well as on the convergence of training loss. As the forth column of Table \ref{tab: ablation} displays, using the second order (third column in the Table \ref{tab: ablation}) renders a slight performance boost over the first order in classification and almost $2\%$ increase in learning to rank task. Figure \ref{tr_losses} displays the effects of first and second order statistics on training loss. The final model (with second order statistics) again shows slightly better convergence on classification yet the difference is more apparent on ranking. As the learning rate is decreased by a rate of $1/2$ per $15$ weak learners, sudden drops are observed in classification loss curve at $15^{th}$ and $30^{th}$ stages in Figure \ref{tr_losses}.

% \begin{figure*}[ht]
% \begin{centering}
% \subfigure[]{\includegraphics[width=0.49\textwidth, trim=0 0 10 0, clip]{figures/hl_exp_higgs_trloss.pdf}}%\label{fig:hl_exp}
% \subfigure[]{\includegraphics[width=0.49\textwidth, trim=0 0 10 0, clip]{figures/hu_exp_higgs_trloss.pdf}}%\label{fig:hu_exp}
% \par\end{centering}
% %\vspace{3mm}
% \caption{Effect of hidden layers and units on classification performance (AUC).}
% \label{fig:hl_hu_exp}
% \end{figure*}

\subsection{Analyzing the effect of hidden layers}
As the literature suggests, boosting algorithms work best with weak learners, thus we utilized a shallow neural network with two hidden layers as a weak predictor for our model. While adding more hidden layers yields stronger predictors, they are not weak learners anymore. To explore this weak learner limit on the number of hidden layers, we assayed weak learners with  1, 2, 3, and 4 hidden layers.
%Figure 2 (a) from Supp. material demonstrates training loss with test AUC scores from GrowNet with 1, 2, 3, 4, and 5 hidden layers. 
Each hidden layer had 16 units. Although weak learners with more hidden layers render better training loss convergence as expected, the overall model starts  saturating on performance and overfitting. Weak learners with 1 and 2 hidden layers attain the best scores, yet the latter one outperforms the former. The worst test AUC score is from the model with 4 hidden layers (See Fig 6 in Supp. material).
\begin{wrapfigure}{r}{5cm}
\vspace{5pt}
\includegraphics[width=0.35\textwidth]{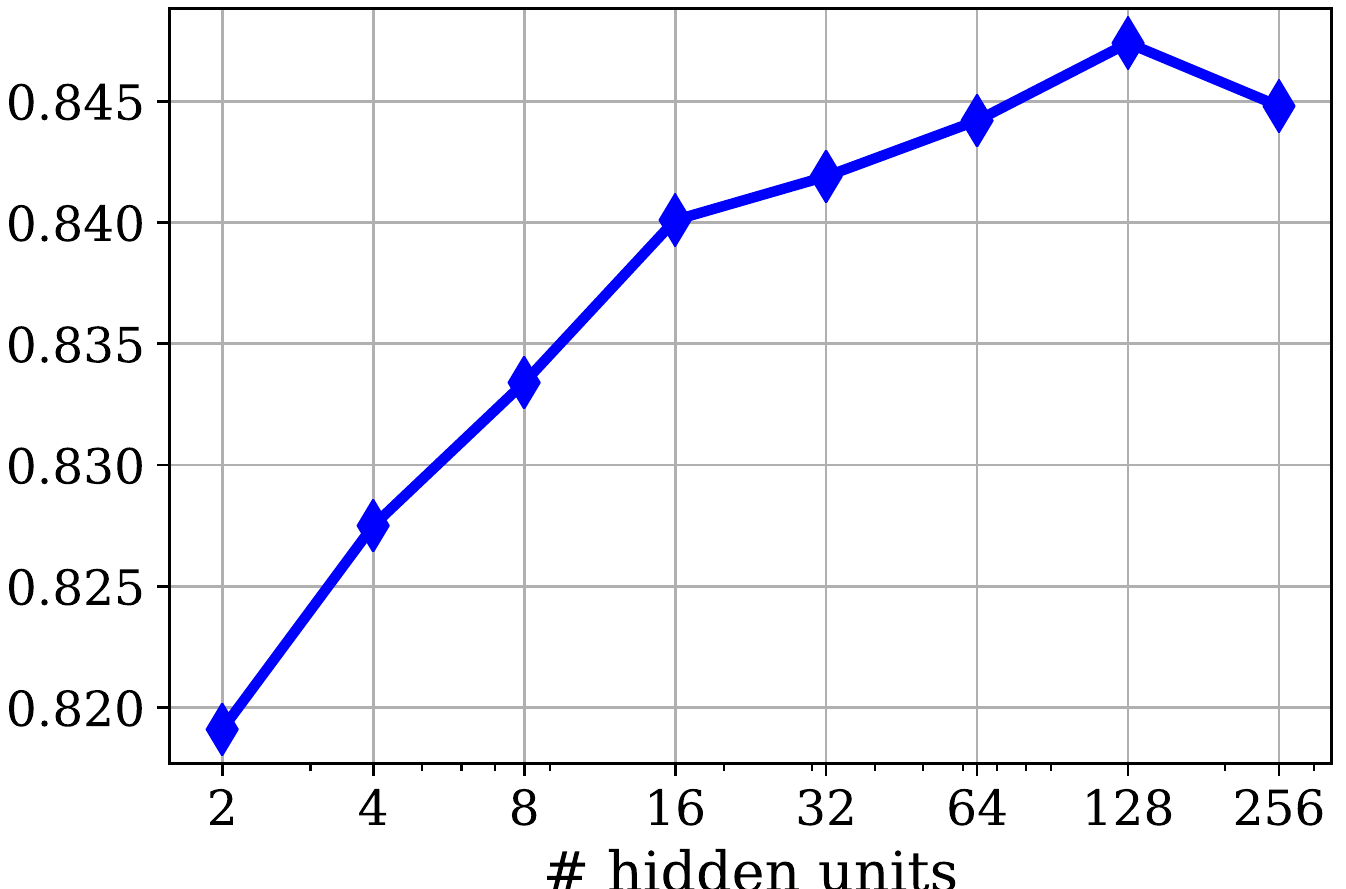}
\caption{Effect of \# neurons on classification performance}
\label{fig:hu_exp}
\vspace{-5pt}
\end{wrapfigure}
 
 Altering the number of hidden units has a lesser effect on performance. To illustrate the impact of hidden layer dimensions, we tested the final model (weak learner with two hidden layers) with various hidden units. Higgs data has 28 features and we tested the model with 2, 4, 8, 16, 32, 64, 128 and 256 hidden units. The smaller the hidden layer dimension is, the less information propagation the weak learners get. On the other hand, having a large number of units also leads to overfitting after a certain point. Figure \ref{fig:hu_exp} displays test AUC scores from this experiment on Higgs 1M data. The highest AUC of $0.8478$ is achieved with 128 units, yet the performance suffers when the number is increased to 256.

\subsection{GrowNet versus DNN}

 One might ask what would happen if we just combine all these shallow networks into one deep neural network. There are a couple of issues with this approach: (1) it is very time-consuming to tune the parameters of the DNN, such as the number of hidden layers, the number of units in each hidden layer, the overall architecture, batch normalization, dropout level, and etc., (2) DNNs require a huge computational power and in general run slower. We compared our model (with 30 weak learners) against DNN with 5, 10 , 20, and 30 hidden-layer configurations. The best DNN (with 10 hidden layers) produced $0.8342$ on Higgs 1M data in 1000 epochs, and each epoch took $11$ seconds. The DNN achieved this score (its best) at epoch 900. GrowNet rendered $0.8401$ AUC on the same configuration with 30 weak learners. The average stage training time, including the corrective step, took 50 seconds. Both models are run on the same machine with NVIDIA Tesla V100 (16GB) GPU. We find that GrowNet has a clear advantage over stacked DNNs on all these aspects.
 
 Further details and illustrations from the ablation study and the code are provided in supplementary material.
\section{Conclusion}

In this work, we propose \textit{GrowNet}, a novel approach to leverage shallow neural networks as ``weak learners" in a gradient boosting framework. This flexible network structure allows us to perform multiple machine learning tasks under a unified framework while incorporating second order statistics,  corrective step and dynamic boost rate to remedy the pitfalls of traditional gradient boosting decision tree. Ablation study is conducted to explore the limits of neural networks as weak learners in the boosting paradigm and analyze the effects of each GrowNet component on the model performance and convergence. We show that the proposed model achieves better performance in regression, classification and learning to rank on multiple datasets, compared to state-of-the-art boosting methods. We further demonstrate that GrowNet is a better alternative to DNNs in these tasks as it yields better performance, requires less training time and is much easier to tune.

\section*{Broader Impact}
GrowNet describes a novel boosting framework, which could be applied to a wide range of tasks and application domains, not limited to classification, regression and learning to rank, which are discussed in this paper. Our research could help improve the performance of these tasks and applications in practice.

While there are general impacts of our work, similar to those of the popular boosting methods~\cite{xgboost, adanet}, we focus on the impact of the ease of employing boosted neural networks in practice. The potential benefits are improved predictions in a lesser amount of time (instead of a longer time to comprise application-specific algorithms) in a wide range of applications. For example, in healthcare, better predictions result in better diagnosis, which could save lives; in social domains, better predictions improve the quality of our lives; the list of tasks and applications, where our method could be adapted to, goes on. While there are undoubtedly benefits, there are many concerns and risks of irresponsible uses. This is even more important in today's environment, where the issues such as bias, discrimination, privacy, etc... increasingly become more serious. Better predictions should not result in bias and discriminate decisions and invasion of privacy. 

To mitigate these risks and concerns, we encourage the research community and policy makers to understand and evaluate the specific impacts of more and more powerful, ready-made Artificial Intelligent algorithms. Here, we need to understand the risks and derive the appropriate policies but should not hinder research activities when there are clearly beneficial implications.

% \begin{ack}
% Use unnumbered first level headings for the acknowledgments. All acknowledgments
% More information about this disclosure can be found at: \url{https://neurips.cc/Conferences/2020/PaperInformation/FundingDisclosure}.

% Do {\bf not} include this section in the anonymized submission, only in the final paper. You can use the \texttt{ack} environment provided in the style file to autmoatically hide this section in the anonymized submission.
% \end{ack}

\medskip

\bibliography{biblio}
\bibliographystyle{icml2019}

\newpage

\appendix

% The \author macro works with any number of authors. There are two commands
% used to separate the names and addresses of multiple authors: \And and \AND.
%
% Using \And between authors leaves it to LaTeX to determine where to break the
% lines. Using \AND forces a line break at that point. So, if LaTeX puts 3 of 4
% authors names on the first line, and the last on the second line, try using
% \AND instead of \And before the third author name.

\section{Additional Related Work}

A few works \cite{boosted_CNN, gb_random_cnn} have also been proposed to directly combine Gradient Boosting with  Convolutional Neural Nets (CNN). The authors of \cite{gb_random_cnn} propose to train gradient boosting machine with  CNN as a base learner by introducing a custom multi-class softmax loss function for a specific scene classification task in the remote sensing domain. The work in \cite{boosted_CNN}, on the other hand, focuses on training each CNN sequentially on the mistakes of the previous networks, similar to Adaboost to perform on solely image classification task. Our method is different from \cite{gb_random_cnn, boosted_CNN} as it is a unified framework to perform various machine learning tasks, such as classification, regression and even learning to rank. Moreover, unlike those two methods, we leveraged a corrective step to update the previously added predictor parameters and achieved a significant performance boost.

\section{Algorithm Pseudocode}

The pseudo-code of training the weak learner is explained in Individual model training part (1) of algorithm \ref{algo:main}. The second part of the algorithm describes the corrective step. The code is available at GitHub page: \url{https://github.com/sbadirli/GrowNet}.
\begin{algorithm}[H]
    \caption{Full GrowNet training}
    \label{algo:main}
\begin{algorithmic}
    \STATE \textbf{Input:} $f_0(x) = log(\frac{n_+}{n_-})$, $\alpha_0$, Training data $\mathcal{D}_{tr}$\\
    \STATE \textbf{Output:} GrowNet $\mathcal{E}$\\
\FOR{$k = 1$ {\bfseries to} $M$}
    \STATE \# Part 1 - Individual model training
    \STATE Initialize model $f_k(x)$
    \STATE Calculate $1^{st}$ order grad.: $g_i = \partial_{\hat{y}_{i}^{(k-1)}} l (y_i, \hat{y}_i^{(k-1)}),$ $\forall x_i \in \mathcal{D}_{tr}$ 
    \STATE Calculate $2^{nd}$ order grad.: $h_i = \partial^{2}_{\hat{y}_{i}^{(k-1)}} l (y_i, \hat{y}_i^{(k-1)}),$ $\forall x_i \in \mathcal{D}_{tr}$ \\
    \STATE Train $f_k(\cdot)$ by least square regression on $\{x_i, -g_i / h_i\}$ \\
    \STATE Add the model $f_k(x)$ into the GrowNet $\mathcal{E}$ \\

    \STATE \# Part 2 - Corrective step\\
    \FOR{$epoch = 1$ {\bfseries to} $T$}
        \STATE Calculate GrowNet output: $\hat{y}_i^{(k)} = \sum_{m=0}^{k} \alpha_m f_m (x_i)$, $\forall x_i \in \mathcal{D}_{tr}$ 
        \STATE Calculate Loss from GrowNet: $\mathcal{L} = \frac{1}{n} \sum_i^{n} l (y_i, \hat{y}_i^{(k)})$
        \STATE Update model $f_m$ parameters through back-propagation  $\forall m \in \{1, ...k\}$\\
        \STATE Update step size $\alpha_k$ through back-propagation
    \ENDFOR
\ENDFOR
\end{algorithmic}
\end{algorithm}

\section{Additional Dataset Statistics}

We evaluate our model on 5 datasets from 3 different tasks. A brief description of these datasets are presented in Table \ref{tab: datasets}. 

We used Higgs Bozon dataset\footnote{\scriptsize{\url{https://archive.ics.uci.edu/ml/datasets/HIGGS}}} for classification. Higgs data is created using Monte Carlo simulations on high energy physics events. It is a binary event classification data with 28 attributes.

For the regression task, 2 datasets from the UCI machine learning repository are selected. The first one is Computed Tomography (CT) slice localization data\footnote{\scriptsize{\url{https://archive.ics.uci.edu/ml/datasets/Relative+location+of+CT+slices+on+axial+axis}}} where the aim is to retrieve the location of CT slices on the axial axis. The data was constructed from a set of $53,500$ CT images that were taken from 74 different patients (43 male, 31 female). 

The second regression dataset is YearPredictionMSD\footnote{\scriptsize{\url{https://archive.ics.uci.edu/ml/datasets/YearPredictionMSD}}} data, a subset of Million Song dataset, from the UCI repository. The goal is to predict the release year of a song from its audio features. The songs are mostly western, commercial tracks ranging from 1922 to 2011, with a peak in the year 2000s.

We choose Yahoo LTRC dataset\footnote{\scriptsize{\url{https://webscope.sandbox.yahoo.com/catalog.php?datatype=c}}} \cite{yahoo_ltrc} for the learning to rank task as it is a well-know benchmark dataset and also is used in the XGBoost paper. This dataset has $20K$ queries, each associated with approximately $22$ documents. Train-test split from the original paper is preserved. The second benchmark ranking dataset we used is MSLR-WEB $10K$\footnote{\scriptsize{\url{http://research.microsoft.com/en-us/projects/mslr/}}}. The dataset contains $10K$ queries, each of which corresponds to a list of $100-200$ documents. 

\begin{table}[t]
\caption{Datasets used in the experiments and their brief description. The second and third columns marked as N, M represent number of samples and feature dimension of the dataset, respectively.}
\label{tab: datasets}
\begin{center}
\begin{tabular}{|l|c|c|c|}
\hline
    Dataset    & N & M & Task\\ \hline
Higgs Bozon & $10M$ & $28$ & Binary classification  \\ \hline
%Criteo ?    & & &  \\\hline
Slice localization & $53K$ & $384$ &  Regression \\\hline
Year prediction & $515K$ & $90$ &  Regression \\ \hline
Yahoo LTRC & $473K$ & $700$ & Learning to rank \\ \hline
MSLR-WEB 10K & $1.2M$ & $136$ & Learning to rank \\ \hline
\end{tabular}
\end{center}
\end{table}

\section{Hyperparameters of GrowNet}
\textbf{Experiment Setup.} All predictive functions added to the model are multilayer perceptrons with two hidden layers. More hidden layers degraded the performance as the model starts overfitting. We generally set the number of  hidden layer units  to roughly a third of, a half of or equal to the input feature dimension. $40$ additive functions were employed in the experiments for all three tasks, and number of weak predictors in test time is chosen by the validation results. From all the experiments, we observe that 30 weak learners are more than enough to get the best results before the model performance saturates. Early stopping or other heuristics can also be incorporated into the model to terminate the training before the model begin to overfit.

The boosting rate is initially set to $1$ and automatically adjusted during corrective step. Depending on the  dataset and the task at hand, it may be initially set to a lower number such as $0.1$. In our experiments, we did not tune or alter the boost rate. 

We trained each predictive function for just $1$ epoch, and the entire model is also trained for $1$ epoch during the corrective step using stochastic gradient descent with the Adam optimizer. The Adam optimizer is run with $l_2$ regularization at a rate of $0.001$. Epoch numbers are increased to 2 for the ranking task as we used larger batch sizes. Increasing the epoch number does not contribute to the performance, and higher numbers cause overfitting. We also performed $2D$ batch normalization for the hidden layers. The batch size for classification was set to 2048 and the learning rate was set to $0.005$.  ReLU was used as the activation function for the penultimate layer, whereas Leaky ReLU was used for the hidden layers. For the ranking task, we replaced ReLU with ReLU6.

The source code is uploaded in a separate file.
\section{XGBoost and AdaNet Tuning}

\subsection{XGBoost Tuning}
For XGBoost, we tuned the main parameters, including the number of trees, learning rate, maximum leaves and $\ell_2$ regularization in the following range:
\begin{itemize}
    \item Number of trees: $\{64, 128, 256, 512, 1000\}$
    \item Learning rate: $\{0.05, 0.1\}$
    \item Maximum number of leaves: $\{128, 256, 512 \}$
    \item $\ell_2$ regularization (\texttt{lambda}): $\{0, 0.2\}$
\end{itemize}

We did not tune XGBoost on Yahoo LTR (ranking task) and Higgs (classification task) datasets as we used the results reported in the original XGBoost paper \cite{xgboost} as is.

\subsection{AdaNet Tuning}
We tuned 3 main parameters for AdaNet: the learning rate, the number of sub-networks and the complexity regularization parameter ($\lambda$) within the following ranges:
\begin{itemize}
    \item Learning rate: $\{0.01, 0.001, 0.0001\}$
    \item AdaNet iterations (\# subnetworks): $\{2, 3, 4\}$
    \item Complexity regularizer $\lambda$: $\{0.01, 0.001, 0.0001\}$
\end{itemize}

The model was first tuned with default mixture weights and $\lambda=0$, as suggested in the authors' Github page\footnote{\url{https://github.com/tensorflow/adanet/blob/master/adanet/examples/tutorials/adanet_objective.ipynb}}. From this experiment, we got the optimal learning rate and number of sub-networks. Then using the learned parameters from the previous setting,  AdaNet is again tuned to learn the mixture weights and regularization complexity parameter $\lambda$. 

The model is trained for $30,000$ epochs, and the number of neurons in layers is set to 512, following the results from AdaNet paper~\cite{adanet}. We also observed that the model with 512 neurons generally renders better performance.

\subsection{Classification on Higgs-1M}
Following the same data split on Higgs data from the XGBoost paper~\cite{xgboost}, we created Higgs-1M data. Table~\ref{tab:higgs_1m_exp} reports the AUC scores on HIggs-1M data from GrowNet, AdaNet and XGBoost. GrowNet renders favorable results compared to XGBoost and $3\%$ increase over AdaNet result. 

\begin{table}[]
    \centering
    \begin{tabular}{|l|c|c|c|}
        \hline
         & GrowNet & AdaNet & XGBoost \\ \hline
      AUC & $\boldsymbol{0.8401}$ & $0.8143$ & $0.8304$  \\ \hline
    \end{tabular}
    \vspace{10pt}
    \caption{Classification results on Higgs-1M data. The scores are in AUC-ROC.}
    \label{tab:higgs_1m_exp}
\end{table}

\section{Additional Illustrations for Ablation Study}

\begin{figure*}[ht]
\begin{centering}
\includegraphics[scale=0.75]{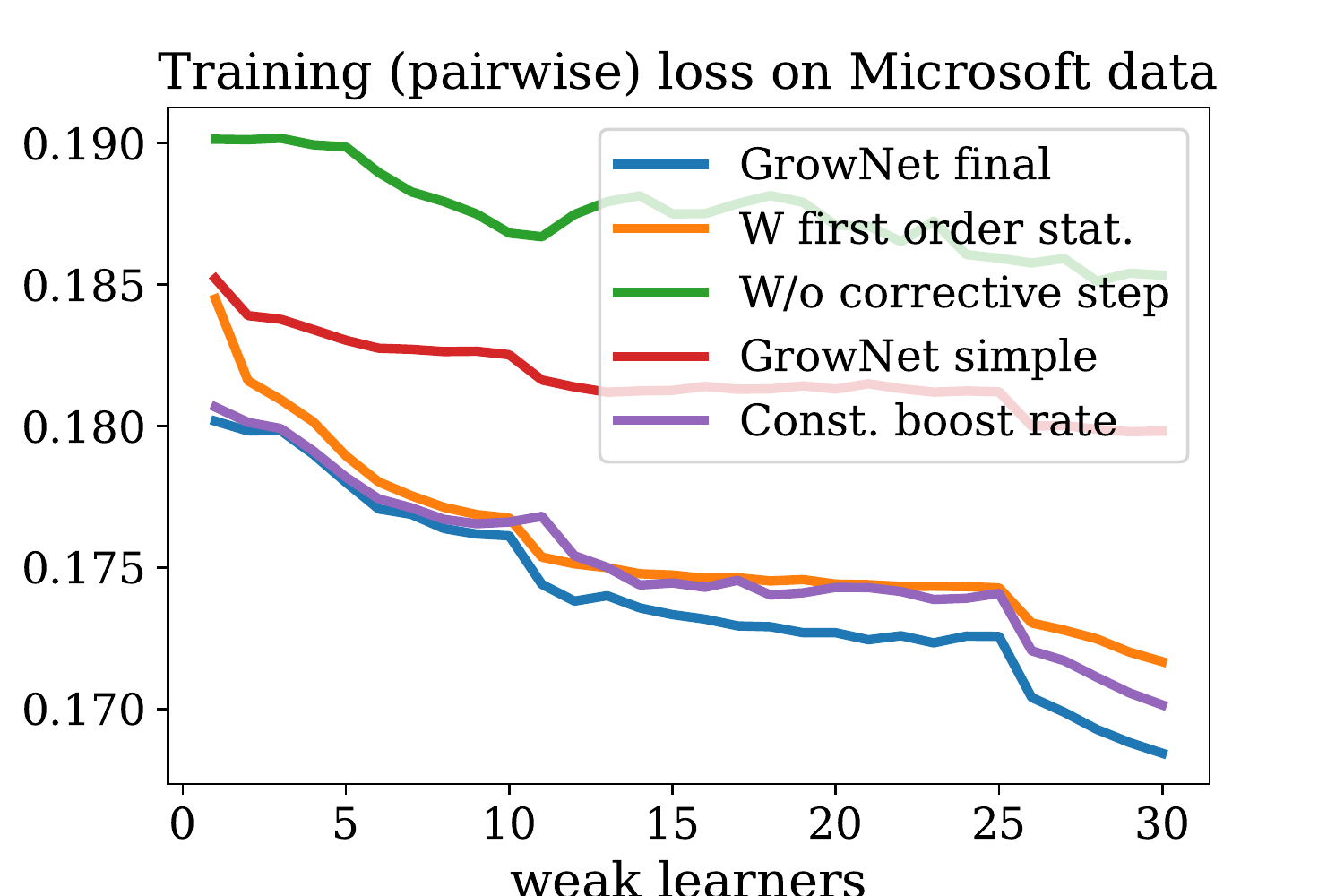}
\par\end{centering}
%\vspace{3mm}
\caption{Training loss visualization for the learning to rank task on MSLR dataset. We used pairwise loss.}
\label{fig:ranking_losses}
\end{figure*}

Analogous to Figure 2 from the main text, Figure~\ref{fig:ranking_losses} presents pairwise losses on Microsoft dataset from the ranking task. 

\textbf{Effect of hidden layers.} Table~\ref{tab:hl_exp} reports the results from the hidden-layer experiment. GrowNet final, employing weak learners with 2 hidden layers, got the best performance (AUC score is $0.8401$). The model with a shallow network of 1 hidden layer as a weak learner obtains better performance (AUC of $0.8336$) once the number of hidden units is increased from 16 to 32. The inverse effect on the model with weak learners of 3 or 4 hidden layers did not work as expected. That is, decreasing the number of neurons in the hidden layers for these predictive functions did not improve much the classification performance.

\begin{table}[]
    \centering
    \begin{tabular}{|l|c|c|c|c|}
    \hline
    & GrowNet 1HL & GrowNet 2HL & GrowNet 3HL & GrowNet 4HL  \\ \hline
     AUC  & $0.8288$ & $0.8401$ & $0.8146$ & $0.7801$ \\ \hline
    \end{tabular}
    \vspace{5pt}
    \caption{Results from hidden layer experiment.}
    \label{tab:hl_exp}
\end{table}

\begin{figure*}[ht]
\begin{centering}
\subfigure[Classification training loss]{\includegraphics[width=0.45\textwidth]{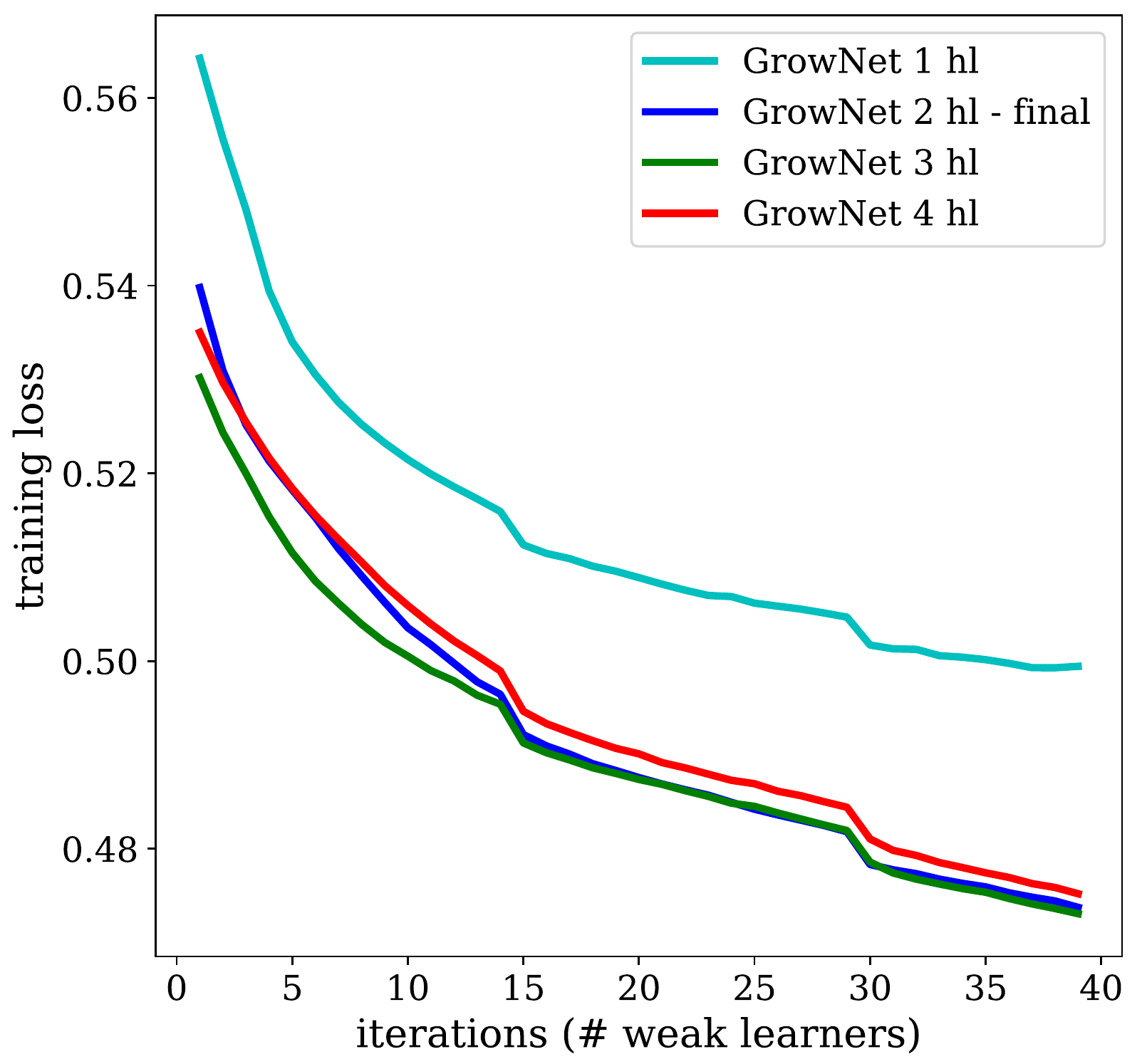}}%\label{fig:hl_exp}
\subfigure[Classification test loss]{\includegraphics[width=0.45\textwidth]{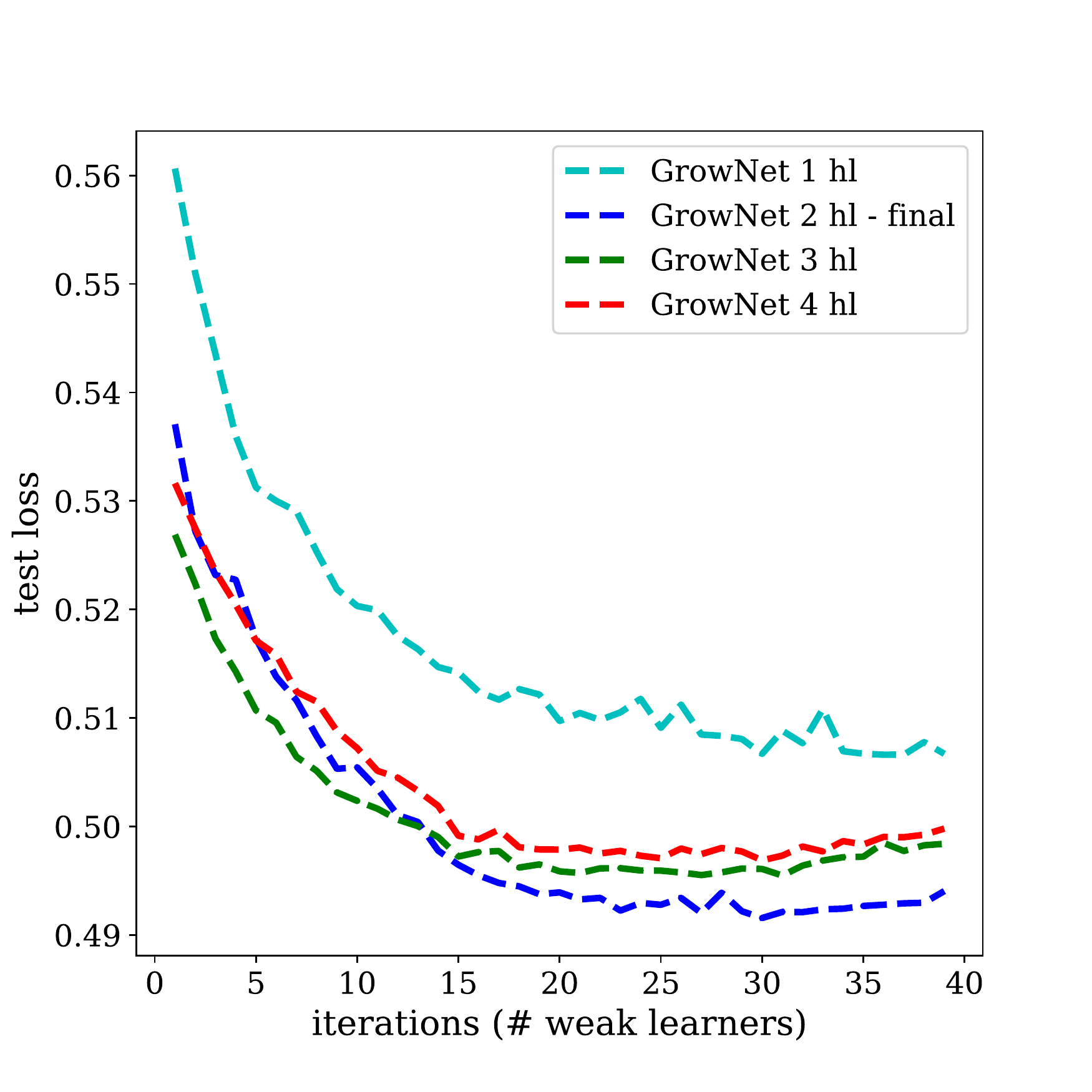}}%\label{fig:hu_exp}
\par\end{centering}
%\vspace{3mm}
\caption{Effect of hidden layers on model training and classification performance (AUC).}
\label{fig:hl_hu_exp}
\end{figure*}

\textbf{Details on DNN versus GrowNet}
Both Deep Neural Network (DNN) models and Grownet are run on the same machine with NVIDIA Tesla V100 (16GB) GPU. 

Unlike GrowNet, DNN performed better with SELU activation functions. We also applied batch normalization on the hidden layers of DNN. Each of DNN models run for 1000 epochs. The results are reported in Table \ref{tab:dnn_vs_grownet}. The best performing DNN model has 10 hidden layers, and each epoch took approximately 12 seconds. The model reaches its best performance after epoch 900. GrowNet shows a clear advantage on both classification performance and training time.

\begin{figure}[h]
    \centering
    \includegraphics[scale=0.5]{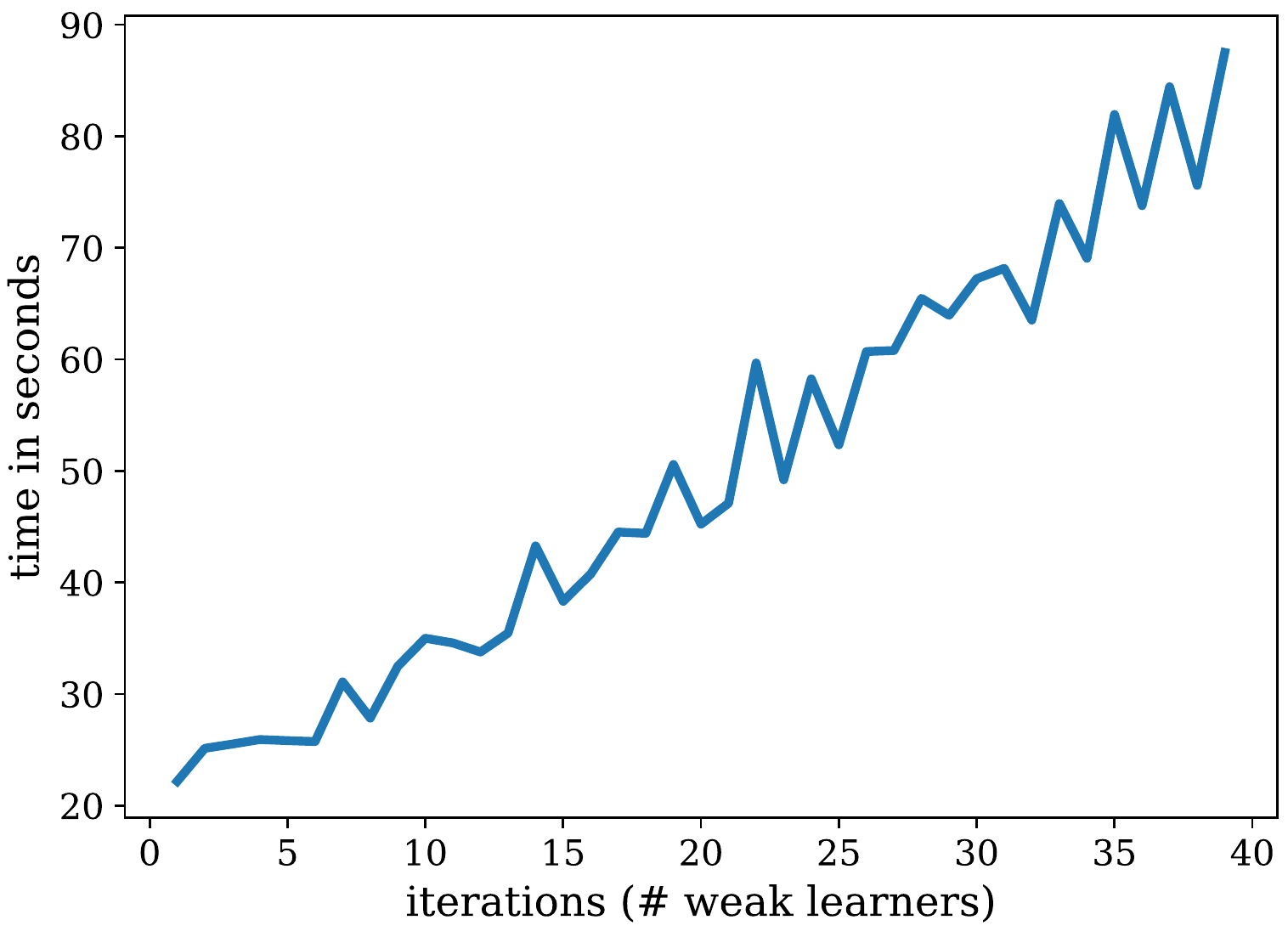}
    \caption{Training time over iterations. As observed, training time is linearly correlated with number of weak learners.}
    \label{fig:grownet_tr_time}
\end{figure}

Both methods, DNN and GrowNet are not fully optimized, thus their training time can slightly be improved. Figure~\ref{fig:grownet_tr_time} displays the training time of GrowNet while adding new weak learners.
DNN with 30 hidden layers are implemented with Dropout(0.3), as without Dropout the model started to overfit immediately after a few epochs. That also explains very close training times of DNN with 20 and 30 layers. 
\begin{table}[h]
    \centering
    \begin{tabular}{|l|c|c|c|c|c|}
    \hline
      Models   & DNN-5 & DNN-10 & DNN-20 & DNN-30 & GrowNet  \\ \hline
    Training time (sec) & $10.2$ & $11.6$ & $15.2$ & $15.0$ &$50.1$ \\ \hline
    AUC & $0.8288$ & $0.8342$ & $0.8338$ & $0.8301$ &$0.8401$ \\ \hline
    \end{tabular}
    \vspace{5pt}
    \caption{Training time and performance comparison between DNN and GrowNet on Higg-1M data. Training time for DNNs are average seconds per epoch and for GrowNet average seconds per stages.}
    \label{tab:dnn_vs_grownet}
\end{table}

\end{document}